# Gesture Recognition Based Mouse Events

Rachit Puri

IM\Web, Multimedia & Services\Web Solutions\Web Engine
Samsung Research India-Bangalore
Bangalore – 560037, India
rachit.puri@samsung.com

## ABSTRACT

*This paper presents the maneuver of mouse pointer and performs various mouse operations such as left click, right click, double click, drag etc using gestures recognition technique. Recognizing gestures is a complex task which involves many aspects such as motion modeling, motion analysis, pattern recognition and machine learning.*

*Keeping all the essential factors in mind a system has been created which recognizes the movement of fingers and various patterns formed by them. Color caps have been used for fingers to distinguish it from the background color such as skin color. Thus recognizing the gestures various mouse events have been performed. The application has been created on MATLAB environment with operating system as windows 7.*

## 1. INTRODUCTION

The research activity on gesture based user interface (UI) has been proliferating in the last decade [1]. The main reason of these technologies becomes more popular is because it can be applied into many different fields easily and efficiently. Especially, gesture recognition easily can be applied to the web services, smart home systems, robot manipulation and games [2]. That is why tracking non-rigid motions from sequential videos have been a great interest to the computer vision community. We grew up interacting with the physical objects around us. How we manipulate these objects in our lives every day, we use gestures not only to interact with objects but to interact with each other and this paper brings us a step closer to Human-object relationship by using gesture recognition technique.

In this research still webcam has been used to recognize the gestures. There is no need of 3D or stereo cameras and above research has also been tested on low cost 1.3 megapixel laptop webcam. My work is dedicated to a simple vision based gesture recognition system that acts like an interface between the user and various computing devices in the dynamic environment. This paper presents the technique to perform numerous mouse operations thus obviating the need of hardware used for interaction between the user and computing device. The same approach can be applied to endless tasks such as browsing images, playing games, changing T.V channels etc.

There is a threshold value for distance (in meters) between the user and camera which can further be varied according to camera's resolution. It means if subject who wants to be recognized with his hand gestures in some environment, subject has to come close to certain fixed distance to the camera [3]. This research was done on 1.3 megapixel webcam with threshold value of 2m.

## 2. ASSUMPTIONS

Following assumptions were taken to make the system work effectively and efficiently:
- Camera is still and continuously capturing the frames.
- The movement of fingers and gestures made are not changed rapidly.
- Color caps are needed on the finger so as to track its movement.

## 3. RELATED WORK

In some passed decades Gesture recognition becomes very influencing term. There were many gesture recognition techniques developed for tracking and recognizing various hand gestures. Each one of them has their pros and cons. The older one is wired technology, in which users need to tie up themselves with the help of wire in order to connect or interface with the computer system. In wired technology user can not freely move in the room as they connected with the computer system via wire and limited with the length of wire.

In [4], structured light was used to acquire 3D depth data; however, skin color was used for segmenting the hand as well, which requires homogeneous and interest points on the surface of the hand using a stereo camera. Motion information obtained from the 3D trajectories of the points was used to augment the range data. One can also create a full 3D reconstruction of the hand surface using multiple views. However, although 3D data contains valuable information that can help eliminate ambiguities due to self-occlusions which are inherent in image-based approaches, an exact, real-time, and robust 3D reconstruction is very difficult. Besides, the additional computational cost hinders its application in real-life systems.

Later on some advanced techniques have been introduced like Image based techniques which require processing of image features like texture etc. If we work with these features of the image for hand gesture recognition the result may vary and could be different as skin tones and texture changes very rapidly from person to person from one continent to other.

To overcome these challenges and promote real time application, gesture recognition technique based on color detection and their relative position with each other has been implemented. The color can also be varied and hence obviating the need of any particular color. The movement as well as mouse events of mouse are very smooth and user is able to select the small menu buttons and icons without any difficulty.

In paper [5], the approach is based on calculation of three combined features of hand shape which are compactness, area and radial distance. Compactness is the ratio of squared perimeter to area of the shape. If compactness of two hand shapes are equal then they would be classified as same, in this way this approach limits the number of gesture pattern that can be classified using these three shape based descriptors.

The algorithm implemented in this paper is divided into seven main steps. First one is selection of RGB value. The second step includes conversion of RGB value to YCbCr. The further steps are region of interests, scale conversion, mirror value and finally the mouse event. The steps have been explained further in detail below.

# 4. GESTURE RECOGNITION TECHNIQUE

The proposed approach is based on detection of number of target colors (region of interest) that triggers the mouse event according to the gesture formed [6]. At the beginning, snapshot is taken while keeping the hand in front of the camera. The user then selects the color cap which will be tracked during gesture formation to perform various mouse events. The color of cap can be varied and needs to be selected when the snapshot of hand is taken as shown in fig1.

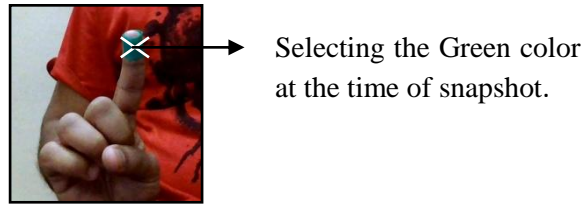

Selecting the Green color at the time of snapshot.

Fig 1. Selecting the color during snapshot

## 4.1 Color Models

A color model is an abstract mathematical model describing the way colors can be represented as tuples of numbers, typically as three or four values or color components. The various color models are: RGB, CYMK, YCbCr etc.

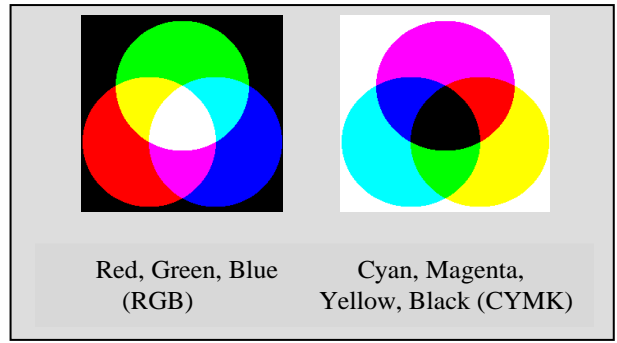

Fig.2 RGB and CYMK color model

The model used in gesture recognition is **YCbCr** [7]. It is a family of color spaces used as a part of the color image pipeline in video and digital photography system. $C_B$ and $C_R$ are the blue-difference and red-difference chroma components. Y is luminance, meaning that light intensity is non-linearly encoded using gamma.

RGB signals are not efficient as a representation for storage and transmission, since they have a lot of mutual redundancy [8]. So, using **YCbCr** model we can separate the luminance factor and hence eliminate its interference with our operation [9]. **YCbCr** is not an absolute color space; it is a way of encoding RGB information. The actual color displayed depends on the actual RGB colorants.

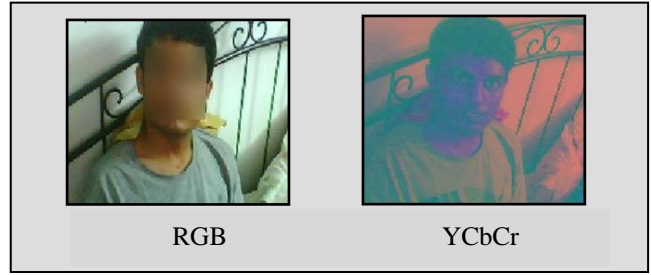

Fig 3. RGB to YCbCr image conversion.

## 5. METHODOLGY

The implementation has been divided into various steps such as selection of RGB, YCbCr conversion, finding region of interest, storing values and last mouse event. Each step has been explained below along with the MATLAB implementation.

**5.1 Selection of RGB**: The first step is to get the RGB value of the color cap which will be tracked during gesture formation to cause mouse events. To perform this step hand with any color cap is kept in front of the camera. The camera will take the snapshot of hand and asks the user to select the color in the snapshot for performing various mouse events. Then user finally selects the color of cap and video starts to track that specific color.

**5.2 YCbCr conversion:** The obtained RGB value in the above step is converted to YCbCr color model using below formula:

$$Y = (0.257 * R) + (0.504 * G) + (0.098 * B) + 16$$
$$Cr = V = (0.439 * R) - (0.368 * G) - (0.071 * B) + 128 \text{ ----- (1)}$$
$$Cb = U = - (0.148 * R) - (0.291 * G) + (0.439 * B) + 128 \text{ ----- (2)}$$

**5.3 Region of Interest:** The next step is to find the region of interest and their relative position that stimulates the mouse events accordingly. To find the region of interest each pixel of captured frame is first converted to Cb' and Cr' using formulas Cb' = im(x, y, 2) and Cr' = im(x, y, 3), where im is captured image; x and y are the co-ordinates.

Now, comparing the Cb' and Cr' with the Cb and Cr obtained in (1) and (2) of color cap with a threshold value of 12 (calculated experimentally), region of interests are calculated.

The threshold value stated above is for green color and can vary slightly depending upon the color and calculated Y value. During experiment it was generally observed that Y value has a direct relation with the threshold value.

**5.4 Storing values:** The number of region of interests and their relative position in terms of pixel value is stored in variables.

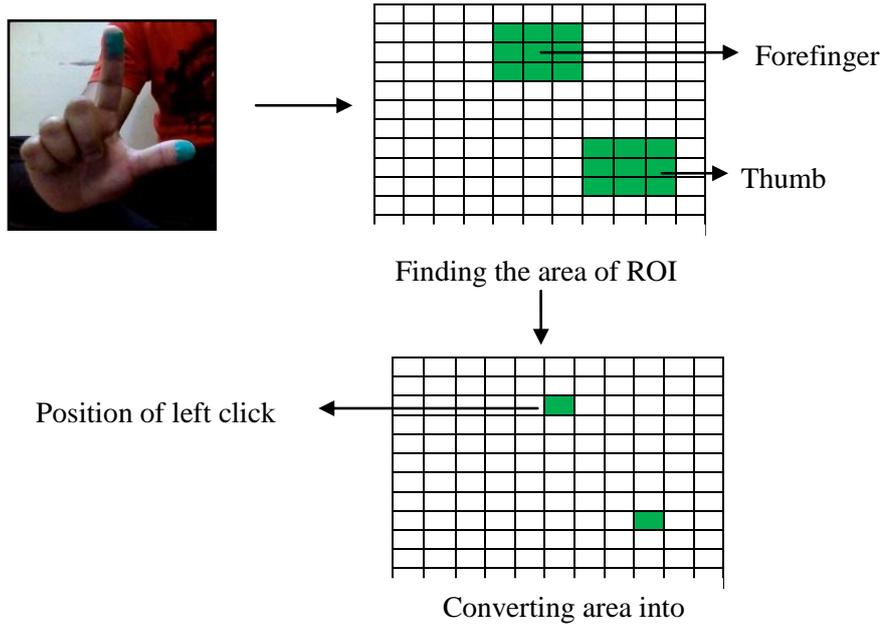

Finding the area of ROI

Position of left click

Converting area into

As shown above the ROI is converted into single pixel value by finding the mid pixel of the shaded region. This single pixel value facilitates the smooth movement of mouse pointer and performs operations equally well.

**5.5 Scale conversion:** The webcam used in the experiment has the resolution of 640 x 480 but the screen resolution is 1600 x 900. Therefore, there is a need of scale conversion.

$$X_s = (1600 \div 640) X_w \quad \text{---- (1)}$$
$$Y_s = (900 \div 480) Y_w \quad \text{----- (2)}$$

$X_s, X_w$: Screen and webcam x pixel value.
$Y_s, Y_w$: Screen and webcam y pixel value.

**5.6 Mirror value:** Once $X_s Y_s$ are obtained, we take the mirror value of $X_s$ as the direction of cursor movement in webcam is opposite to the hand movement. The axis of webcam and screen are coincided before the mirror value of $X_s$ is taken.

**5.7 Mouse Event:** The mouse event such as cursor movement, left click, right click, drag etc are performed according to the number of region of interests and their relative position in each captured frame.

# 6. MATLAB IMPLEMENTATION

```
vid = videoinput ('winvideo', 1,'YUY2_320x240');
set (vid,'TriggerRepeat', Inf);
vid.FrameGrabinterval = 1;
set (vid,'ReturnedColorSpace','rgb');
start (vid);
pause (2);
g = getsnapshot (vid);
p = impixel (g);
disp (p);
close;
yo = (0.257*p (1)) + (0.504*p (2)) + (0.098*p (3)) + 16;
disp (yo);
cbo = (-0.148*p (1)) + (-0.291*p (2)) + (0.439*p (3)) + 128;
disp (cbo);
cro = (0.439*p (1)) + (-0.368*p (2)) + (-0.071*p (3)) + 128;
pause (1);
Now, video is started as {while (vid.FramesAcquired <= 1000000)} and depending
```

## 6.1 Various Mouse Events

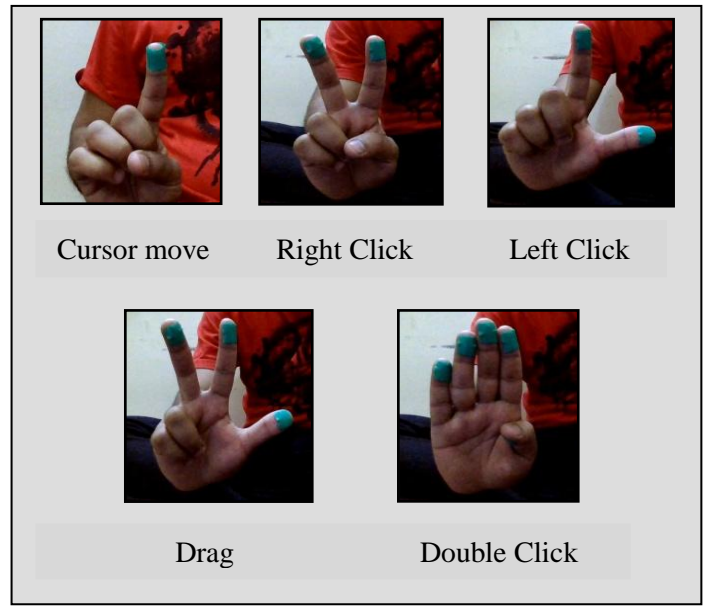

Fig 4. Gesture Vocabulary for mouse events.

## 6.2 Data Flow

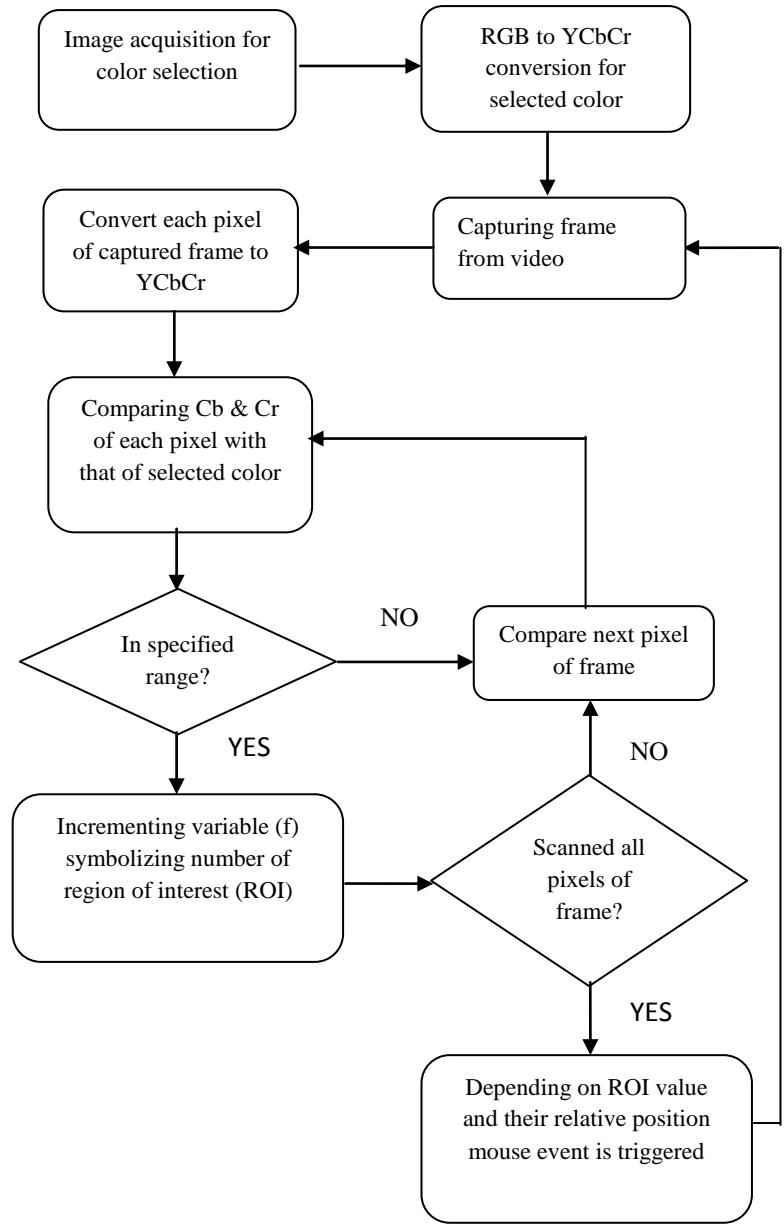

F = 1: Move the cursor to desired position.
F = 2: a) Pixel difference of 2 ROI > threshold value;
    Perform left click.
  b) Pixel difference of 2 ROI < threshold value;
    Perform right click.
F = 3: Perform drag action.
F = 4: Double click.

## 7. EXPERIMENTS AND RESULTS

The results obtained by the experiments for various mouse events such as cursor movement, left click, right click, double click, drag etc has been shown in below graph. The distance between the user and camera was kept fixed as 1m for below experiment.

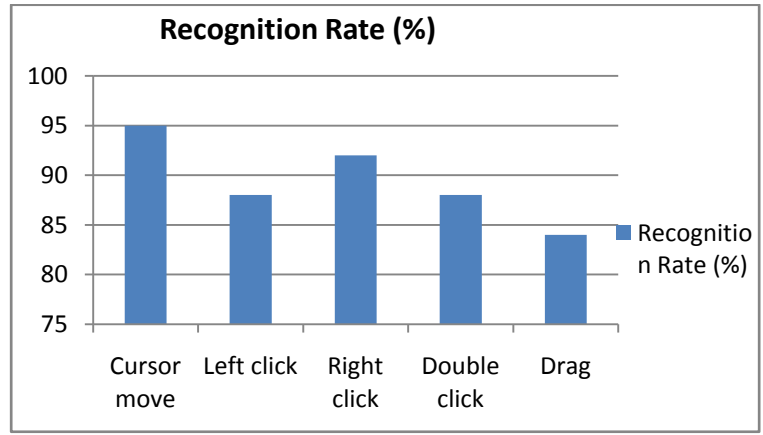

Fig 5. Recognition rate v/s mouse events.

The higher the percentage of recognition rate the more likely the gesture is going to be recognized e.g. cursor move may fail 5% of time i.e. there may be a delay in moving the cursor to its new position.

Below graph shows the response rate of cursor movement when distance between user and camera is varied from 0.3m till 2.4m. The smoothness in cursor movement decreases when the distance from the webcam is increased. For both the experiments 1.3 megapixel laptop webcam was used.

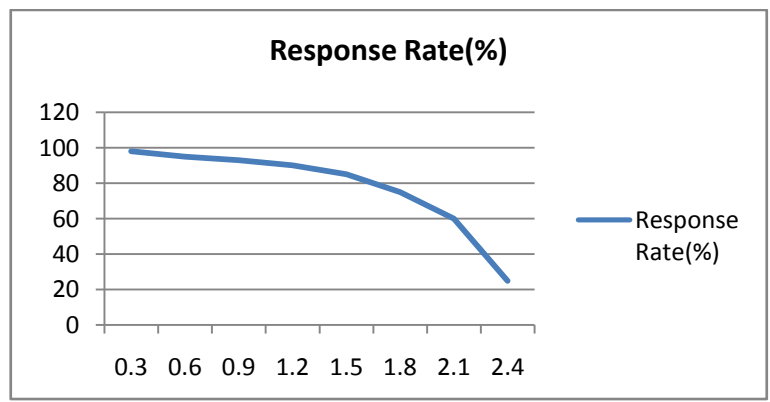

Fig 6. Recognition rate v/s Distance

## 8. CONCLUSION AND FUTURE WORK

Gesture recognition technique shows the positive signs for performing various mouse events. The use of YCbCr color model in research extricates the dependency on light intensity during experiment.

This technique is very useful for crippled people, at the time of presentation using IP cameras, controlling TV channels etc and can be applied to smartphones for numerous operations. Future works will include better methods for implementing mouse events and reducing the lag to almost zero during cursor movement. More features such as zoom in, zoom out, shut down etc will be implemented.

**Author biography**

Rachit Puri received B.Tech degree from Thapar University, Patiala in Electronics and Communication Engineering. He has presented one National paper "Prompt Indian Coin Recognition with Rotation Invariance Using Image Subtraction Technique". Attended two workshops and 5 seminars related to image, sound and video processing. Presently he is working with Samsung India Research – Bangalore.

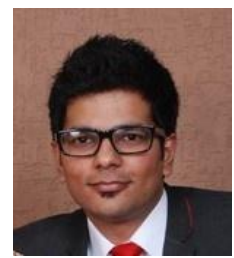